\documentclass[10pt,twocolumn,letterpaper]{article}

\usepackage{iccv}
\usepackage{cite}
\usepackage{times}
\usepackage{epsfig}
\usepackage{graphicx}
\usepackage{amsmath}
\usepackage{amssymb}
\usepackage{algorithm}
\usepackage{algorithmic}
\usepackage{subfigure}
\usepackage{booktabs}
\usepackage{tikz}
\usepackage{makecell}
\usepackage{multirow}
\usepackage{authblk}
\usepackage[T1]{fontenc}

\DeclareUnicodeCharacter{2212}{-}

\usepackage[breaklinks=true,bookmarks=false]{hyperref}
\newcommand\nnfootnote[1]{%
  \begin{NoHyper}
  \renewcommand\thefootnote{}\footnote{#1}%
  \addtocounter{footnote}{-1}%
  \end{NoHyper}
}
\iccvfinalcopy % *** Uncomment this line for the final submission

\def\iccvPaperID{****} % *** Enter the ICCV Paper ID here

\ificcvfinal\fi

\begin{document}

%%%%%%%%% TITLE
\title{Towards Robust Natural-Looking Mammography Lesion Synthesis on Ipsilateral Dual-Views Breast Cancer Analysis}

% \author{Huy T. Nguyen\inst{1} \and
% Hien Q. Kha\inst{2}\thanks{Huy T. Nguyen and Hien Q. Kha equally contribute to this paper.} \and
% Toan N. T. Truong\inst{1, 4} \and Thinh B. Lam\inst{1} \and \ Hung B. Ngo \and Vinh Q. Dinh\inst{3} \and Nguyen Quoc Khanh Le \inst{3}}

% \author{Huy T. Nguyen\\
% \and

\author[1,3]{Thanh-Huy Nguyen}
\author[2,3]{Quang Hien Kha}
\author[4]{Thai Ngoc Toan Truong}
\author[3]{Ba Thinh Lam}
\author[5]{Ba Hung Ngo}
\author[6]{Quang Vinh Dinh}
\author[2, 7, 8]{Nguyen Quoc Khanh Le}
\affil[1]{Department of Biomedical Engineering, National Cheng Kung University, Taiwan}
\affil[2]{International Ph.D. Program in Medicine, College of Medicine, Taipei Medical University, Taiwan}
\affil[3]{Saigon Precision Medicine Research Center (SAIGONMEC), Vietnam}
\affil[4]{Ho Chi Minh City International University, Vietnam}
\affil[5]{Graduate School of Data Science, Chonnam National University, Korea}
\affil[6]{Vietnamese-German University, Vietnam}
\affil[7]{Research Center for Artificial Intelligence in Medicine, Taipei Medical University, Taiwan}
\affil[8]{Professional Master Program in Artificial Intelligence in Medicine, Taiwan}

\maketitle
\def\thefootnote{*}\nnfootnote{*Thanh-Huy Nguyen and Quang Hien Kha have equal contributions.}
\ificcvfinal\thispagestyle{empty}\fi

%%%%%%%%% ABSTRACT
\begin{abstract}
   In recent years, many mammographic image analysis methods have been introduced for improving cancer classification tasks. Two major issues of mammogram classification tasks are leveraging multi-view mammographic information and class-imbalance handling. In the first problem, many multi-view methods have been released for concatenating features of two or more views for the training and inference stage. Having said that, most multi-view existing methods are not explainable in the meaning of feature fusion, and treat many views equally for diagnosing. Our work aims to propose a simple but novel method for enhancing examined view (main view) by leveraging low-level feature information from the auxiliary view (ipsilateral view) before learning the high-level feature that contains the cancerous features. For the second issue, we also propose a simple but novel malignant mammogram synthesis framework for upsampling minor class samples. Our easy-to-implement and no-training framework has eliminated the current limitation of the CutMix algorithm which are unreliable synthesized images with random pasted patches, hard-contour problems, and domain shift problems. Our results on VinDr-Mammo and CMMD datasets show the effectiveness of our two new frameworks for both multi-view training and synthesizing mammographic images, outperforming the previous conventional methods in our experimental settings. 
\end{abstract}

%%%%%%%%% BODY TEXT
\section{Introduction}

Breast cancer has one of the highest rates of mortality and incidence among women worldwide, making it one of the most common cancers to cause death. Cancer detection, in particular at the early stage, must be crucial in screening mammogram exams. Both the craniocaudal (CC) view and the mediolateral oblique (MLO) view, which are top-down and side views of the breast, respectively, can be used to classify each patient's breasts. Radiologists frequently examine both views of the same breast (ipsilateral views) and the same view of both breasts (bilateral views) to make a sound, intuitive decision.  

\begin{figure}[t]
	\centering
	\includegraphics[width=0.7\linewidth]{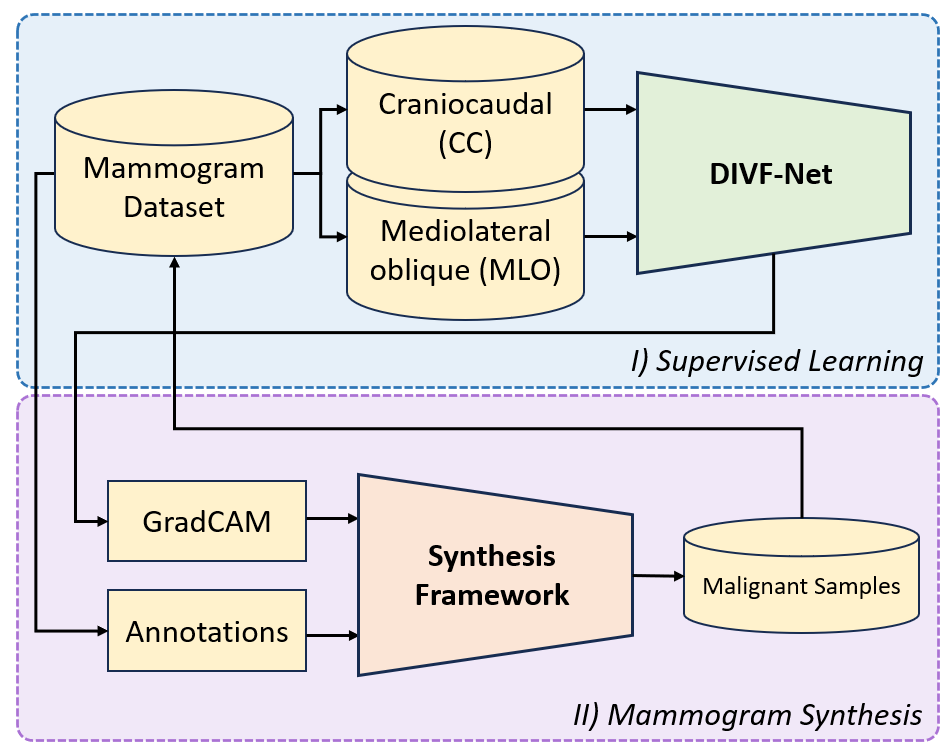}
	\caption{Our proposed pipeline for training and synthesizing mammographic images. Two stages are the supervised training on ipsilateral views mammograms. and synthesis framework that takes the saliency map and region malignant annotations.}
	\label{fig: 0}
 \vskip -0.5in
\end{figure}

Based on that, prior works nowadays can be classified into various groups: ipsilateral-only based, bilateral-only based, and ipsilateral-bilateral combination. Recent papers expose bilateral-only based, Liu et al. \cite{liu2019unilateral} enhanced mammogram mass detection using a contrasted bilateral network (CBN).
Furthermore, Zhao et al. \cite{zhao2021bascnet} used a well-known attention module between adaptive spatial and channel that yields the categorization. In contrast, those strategies still struggle with the conflict between the two breast sides that cause noise in the model because one patient might have the disease on one breast while another does not. Another group is the ipsilateral-bilateral combination approach used three views or four views as the inputs which create a full overview of breasts.  Liu et al. \cite{liu2021act,liu2020cross} achieved this by proposing a remarkable  Anatomy-aware Graph Convolutional Network (AGN) that relies on the mass shape and region to construct the graphical correspondence among different mammographic views. Although the performance of these models is noteworthy, they require massive computational, thus might be hardly embedded in hospital facilities. Continuously, Nguyen et al. \cite{nguyen2022novel} proposed four views input, independently, each view will be learned to extract features and then fed into Light-GBM\cite{ke2017lightgbm} classifier for prediction. Afterward, the result operates the max function between ipsilateral view sides, which can be inaccurate and lead to a poor learning process.

Furthermore, mammogram synthesis and augmentation techniques are also one of the most promising approaches for handling class imbalance. MixUp \cite{zhang2017mixup} reduces the proportion of informative pixels of two images to produce a new image that shows impressive results in many medical image applications. Similar to MixUp, the CutMix \cite{yun2019cutmix} algorithm generates a simple augmentation methodology that replaces the patch of two images together. However, both of these methods might cause a conflict in the label because the random of choosing a patch in the mammographic image can create two different classes in the same image. About MixUp, the algorithm itself uses the image-level mixing between two images without semantic label preserving for mammographic cancer classification. Besides, the region generated from CutMix is random, that might or might not contain the cancerous information when conducting copy-paste. CutMix also might create new untrustworthy samples due to the solid rectangle's boundary of pasted patches and the difference in style space.

To take full advantage, we propose a Dual Ipsilateral Views Fusion Network (DIVF-Net) for mammographic image classification. This network can be separated into three parts: Low-Level Feature Blocks, Features Fusion Blocks, and High-Level Feature Blocks. Our network can leverage low-level information such as the shape, contour, and density of the breasts. The DIVF-Net combines two low-level features for extracting the relevant information before using it for enhancing the main view feature. The high-level information part of DIVF-Net aims to focus on the lesions that highly contain semantic information for cancer classification. Additionally, a Malignant Lesions Synthesis Framework also is proposed in this paper which overcomes the current limitations of CutMix and MixUp algorithms. It includes three stages: Region Selection, Domain Adaptation, and Soft Contour Transformation. The framework carefully picks the radiologist-annotated region for replacing the benign-information-contained region. The rest of the framework aims to close the gap of different between source and target patches before replacing it with a gradient-contour MixUp algorithm.

In summary, the main contributions of our work are as follows: 
\begin{itemize}
\item A novel multi-view network DIVF-Net with two types of fusion operations that leverages information on both CC and MLO views for accurate cancer classification.
\item A new robust mammogram synthesis framework that replaces the benign to malignancy region with an informative region. The created patches are also being smoothed and Fourier-adapted before replacing the indicated regions.
% \item A new robust mammogram synthesis framework that selects informative regions to conduct CutMix. The created patches are also being smoothed and Fourier-adapted before replacing the indicated regions.
\item Experimental results and ablation studies based on a combination of these two show the robustness and generalizability on multiple fusion settings and datasets.
\end{itemize}

\begin{figure*}[t]
	\centering
	\includegraphics[width=0.85\linewidth]{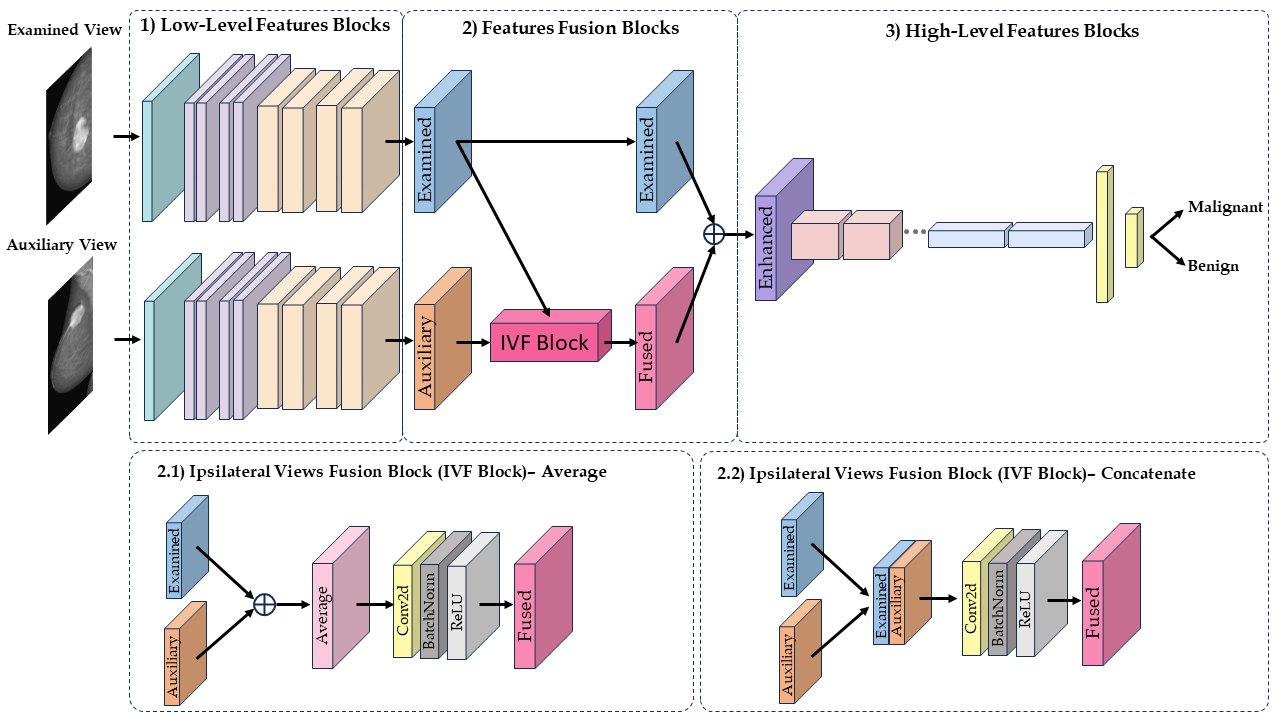}
	\caption{Dual Ipsilateral Views Fusion Network (DIVF-Net) for Mammographic Cancer Diagnosis. This framework consists of three stages: Low-Level Features Blocks (left-top), Features Fusion Blocks (middle-top), and High-Level Features Blocks(right-top). In Ipsilateral Views Fusion (IVF) Block, there are two operations used to fuse both examined features and auxiliary features: average and concatenate. After going through IVF Block, the fused features combine with examined features to improve the performance.}
	\label{fig: 1}
\end{figure*}

\section{Related Work}
\textbf{Multi-view Network:} Compared with 2D views, 3D objects have much more knowledge to guide the model, which is described in visual understanding \cite{qi2016volumetric, yifan2018gvcnn} and stereo vision \cite{chang2018cvpr,xiaozhi20183d, nguyen2018local}. In visual understanding, they set several cameras around a target object to model region-to-region and views from various angles. Each view is embedded in a shared weight Convolutional Neural Network (CNN). In stereo vision, two cameras are placed closely. This approach is mainly used in self-driving cars, which manipulate the depth estimation via disparity map fusion. The depth estimation helps the system knows the closeness to itself the straight object to immediately avoid the car collapse and keep a safe distance. Multi-view-based approaches \cite{su2015multiview, nguyen2022novel} collect features from various 2D views to represent the 3D object. First, they fed each view into a feature extractor to learn the appropriate embedding feature. Then, they proposed their work to significantly fuse all of them for 3D representation. 

Inspired by that, mammographic screening also has a differentiated imaging process that is efficient to represent 3D objects. \emph{Wu et al} \cite{wu2020deep} proposed the four views mammogram network to predict the malignant or not malignant classification. They aggregate between the bilateral views at the first stage and then the softmax layer. Finally, they presented four strategies with a combination of several layers. \emph{Khan et al} \cite{khan2019multiview} enhanced the way mammogram image preprocessing and decreased the computational complexity in the backbone. They extract directly a mass via augmented ROIs and modify a small VGGNet-like architecture used for the feature extraction stage. In general, ipsilateral views consist of CC and MLO views of the same breast side. This advantage in extracting rich information for 3D medical image analysis. Thus, fusing the ipsilateral views increases the global features in fusion operation beside the local features from individual views.

\textbf{Medical Image Synthesis/Augmentation:} Augmentation is one of the most fundamental procedures for synthesizing training data for further generalizability. Existing works on data augmentation \cite{zhang2017mixup, yun2019cutmix, lee2020mixout} synthesize two images into soft images. Thus, the generated new training images direct the model to concentrate more on shape than texture, which improves classification and object identification performances. CutOut \cite{devries2017improved} revivals the object occlusion, which is a common issue in many computer vision tasks. It randomly chooses one defined size patch to remove. While CutMix \cite{yun2019cutmix} replaces the binary mask with another image and mixes the label via the combination ratio. Mixup \cite{zhang2017mixup} sampling from the mixup vicinal distribution produces virtual feature-target vectors.

% there are many robust and promising deep-learning-based medical image synthesis frameworks, such as Generative Adversarial Networks (GAN) \cite{goodfellow2020gan}
% of up to 1280x1024 pixels and achieving positive outcomes
In recent years, Generative Adversarial Networks (GAN) \cite{goodfellow2020gan} become a well-known deep-learning-based medical image synthesis framework. For the synthesis of mammograms, Dimitrios Korkinof et al \cite{korkinof2019pgan}. employ a progressive GAN (PGGAN), achieving high resolutions and positive outcomes when comparing the low-level pixel distributions of real and artificial images. Rui Man et al.'s research \cite{rui2020anogan} focuses on creating synthetic samples, but in this instance, they create patches of a histopathological image. This AnoGAN (Anomaly Detection GAN) has many benefits for teaching classification systems. Xiangyuan Ma et al.'s \cite{Ma_2020} research focuses on creating samples of mammogram lesion segmentation masks. This enables overcoming image labeling, one of the most difficult tasks involved in dataset construction. Having said that, the biggest concern of GAN-based approaches is the realism and trustworthiness of synthesized samples. It may not be practical in real-world applications when using synthesized mammograms for training and testing.

\section{Methodology}
\subsection{Dual Ipsilateral Views Fusion Network} \label{divf}
\begin{figure*}[t]
	\centering
	\includegraphics[width=0.9\linewidth]{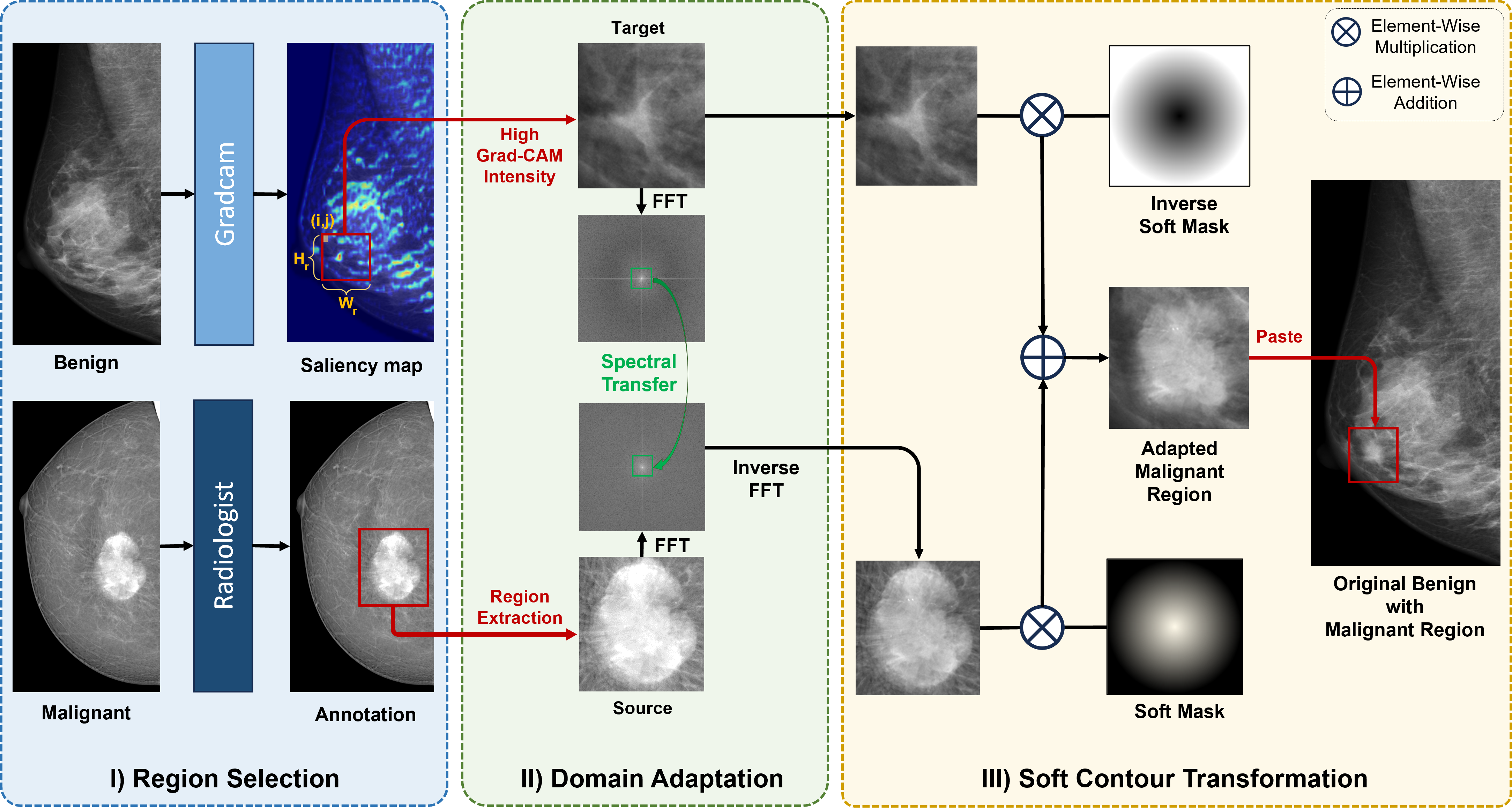}
	\caption{Proposed Soft-Adapted Malignancy Synthesis Framework consists of three phases: 1) Region Selection, extracting the important region (left); 2) Domain Adaptation, adapting target style to the source image (middle); 3) Soft Contour Transformation, smoothing the target region with an inverse soft mask and the transformed source region with a soft mask (right).}
	\label{fig: 2}
\end{figure*}

This work aims to exploit the dual-view mammograms of the same breast using a new proposed network, DIVF-Net. Our network takes two ipsilateral views (CC and MLO) of a single breast to assess the cancerous. For each patient, the model takes one view of the breast as an examined view, and the other is an auxiliary view to support. As shown in Fig.~\ref{fig: 1}, both examined view and auxiliary view are fed into Low-Level Features Blocks (the first half of the popular backbone like ResNet \cite{he2016deep}). Then, the output features of these views are combined by IVF Block. The IVF Block includes 4 components: Aggregation mechanism, 2D convolutional layer, batch normalization layer~\cite{ioffe2015batch}, and ReLU activation function~\cite{agarap2019deep}. In the aggregation part, there are two ways to combine two feature maps: Average and Concatenation, shown in 2.1 and 2.2 of Fig.~\ref{fig: 1}. 

For average aggregation, the output feature takes two feature maps to compute using element-wise average before being fed to the other three components in IVF Block.

For concatenate aggregation, the output feature is a depth stack of two input feature maps before the convolutional layer takes a two-dimension-depth feature map to get the one-dimension-depth feature map. The batch normalization layer and ReLU activation function remain the same as average aggregation for normalizing the input features.

To enhance the examined view with informative information, the output feature map of IVF Block and examined view feature map are combined by element-wise addition. The High-Level Features Blocks (the last half of the backbone) take the enhanced feature maps to learn the high-level information such as abnormalities. Subsequently, we feed it into fully connected layers, followed by a softmax layer, to get the final output binary classification.

This framework's concept is based on how radiologists examine mammograms for diagnosis. Instead of treating two ipsilateral views equally for cancer diagnosis, the model seeks to distinguish one as the primary view and the other as a support view. As shown in Part 2 of Fig.~\ref{fig: 1}, the examined view feature and fused feature play important roles in classifying breast cancer. The examined view is the radiologist's main focus, which is kept the same. On the other hand, the auxiliary view along with examined view is for comparing these two to having more perspectives.

\subsection{Malignant Lesions Synthesis Framework}

\begin{figure*}[ht]
	\centering
	\includegraphics[width=0.85\linewidth]{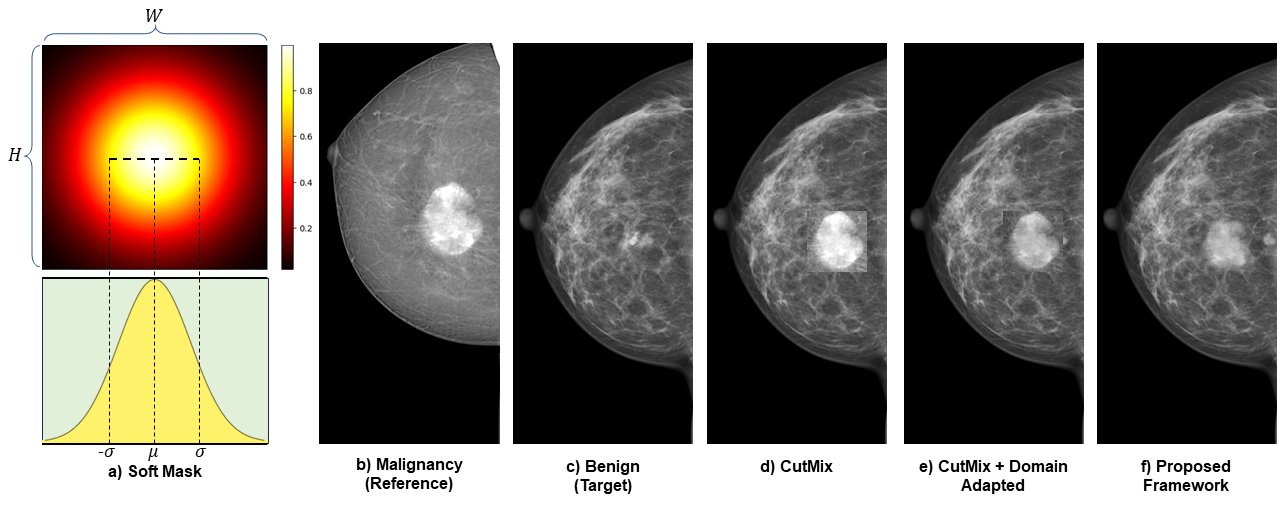}
	\caption{The synthesis mammogram image with various algorithms. a) Soft Mask: following the Gaussian distribution (bottom) to generate the blending masks (top), b) Reference image contains malignancy mass and c) Target image will be added malignancy mass, d) the hard region synthesis CutMix algorithm image, e) the middle smooth region synthesis image with CutMix and Domain Adapted algorithms, f) Our proposed Soft-Adapted Malignancy Synthesis image.}
	\label{fig: 3}
\end{figure*}

Inherited from the previous successful use of domain adaptation on mammogram classification \cite{wang2021deep}, and mammogram detection tasks \cite{nguyen2023context}, 
we proposed a novel framework to create the natural-looking malignant findings synthesis framework. The framework includes three stages:
\begin{enumerate}
\item We first propose a way to select the important region from the benign breast by getting a saliency map from warm-up pre-trained supervised learning. Then, the region with a high-intensity score was replaced by a malignant region that was annotated by radiologists.

\item Secondly, To solve the domain shift issue brought on by breast density or device differences, we conduct the style transfer from the source region to the target region based on Fourier Domain Adaptation\cite{yang2020fda}.

\item Finally, to make the malignant lesions naturally mix with the destination region, we propose a soft contour mask and its inverse to combine source and target regions before pasting to a region of the benign sample.
\end{enumerate}
In the region selection part, the supervised training for warming up is conducted before getting a saliency map. Grad-CAM \cite{selvaraju2017grad} uses the gradient information flowing into the last convolutional layer of the CNN to assign importance values to each neuron for a particular decision of interest. Mathematically, with given class $c$, the saliency map from Grad-CAM $L_{GC}^c \in {R^{H \times W}}$ of height $H$ and width $W$ is obtained by computing the gradient of the class $c$ score and $y^c$ with respect to feature map activations $A^k$ of the convolutional layer (denotes by  $\frac{{\partial {y^c}}}{{\partial {A^k}}}$). To obtain the neuron importance weights (called $a^c_k$), these gradients are global-average-pooled over the width and height dimensions (indexed by i and j, respectively):

\begin{equation}
    a_k^c = \frac{1}{Z}\sum\limits_i {\sum\limits_j {\frac{{\partial {y^c}}}{{\partial A_{ij}^k}}} }.
\end{equation}
% \[a_k^c = \frac{1}{Z}\sum\limits_i {\sum\limits_j {\frac{{\partial {y^c}}}{{\partial A_{ij}^k}}} } \].
    
To obtain $L_{GC}^c$, we perform a weighted combination of forward activation maps followed by a ReLU:

\begin{equation}
    L_{GC}^c = ReLU\left( {\sum\limits_k {a_k^c{A^k}} } \right).
\end{equation}
% \[L_{GC}^c = ReLU\left( {\sum\limits_k {a_k^c{A^k}} } \right)\].

Based on radiologists' malignant abnormalities annotations with width $W_{r}$ and height $H_{r}$, coordinates $(i, j)$ are set as top-left corner starting coordinates. We calculate the bottom-right saliency area value with beginning coordinates $i, j$ and the shape of regions $H_{r}, W_{r}$. Region value of class-discriminative localization map from Grad-CAM, called $I_{region}\left( {L_{GC}, i, j,H_{r},W_{r}} \right)$, is defined as:

\begin{equation}
\label{eqn:cpt_region}
    {I_{region}}\left( {L_{GC},i,j,H_r,W_r} \right) = \sum\limits_{m=i}^{H_{r} + i - 1} {\sum\limits_{n=j}^{W_{r} + j - 1} {{L_{GC}}\left( {m,n} \right)} }.
\end{equation}
% \[{I_{region}}\left( {L_{GC},i,j,H_r,W_r} \right) = \sum\limits_{m=i}^{H_{r} + i - 1} {\sum\limits_{n=j}^{W_{r} + j - 1} {{L_{GC}}\left( {m,n} \right)} } \].

For selecting a wanted region given class c as a pasting destination, we compute the values of regions and find the highest class-discriminative patch as below:
\begin{equation}
    {I_{region}^{*}} = {I_{region}}\left( {L_{GC},i^{*},j^{*},H_{r},W_{r}} \right),
\end{equation}
% \[{I_{region}^{*}} = {I_{region}}\left( {L_{GC},i^{*},j^{*},H_{r},W_{r}} \right) \],
whereas $i^{*},j^{*}$ are computed by:

\begin{equation}
    {i^{*},j^{*}} = \mathop {\arg \max }\limits_{i,j}{{I_{region}}\left( {L_{GC},i,j,H_{r},W_{r}} \right)}.
\end{equation}
% \[{i^{*},j^{*}} = \mathop {\arg \max }\limits_{i,j}{{I_{region}}\left( {L_{GC},i,j,H_{r},W_{r}} \right)} \].

Using $i^{*}, j^{*}$ with $H_{r}, W_{r}$, we can get the patch containing the benign information for mixing. The detailed pseudo-code is described in Algorithm~\ref{alg_select} below.

\begin{algorithm}
    \caption{High class-discriminative Region Selection}\label{alg_select}
    \begin{algorithmic}
    \STATE \textbf{Input:} $H, W, H_{r}, W_{r}$.
        \FOR{$i = 1$ to $H-H_{r}+1$}
            \FOR{$j = 1$ to $W-W_{r}+1$}
                \STATE Calculate $I_{region}$ using Eq.\ref{eqn:cpt_region} \{Compute accumulative intensity of Region saliency map.\}
                \IF{$I_{region}^{*} < I_{region}$}
                    \STATE $I_{region}^{*} \leftarrow I_{region}$ \{Update the biggest intensity of region.\}
                    \STATE $(i^{*}, j^{*}) \leftarrow (i, j)$ \{Update coordinate of the biggest intensity of region.\}
                \ENDIF
            \ENDFOR 
        \ENDFOR
    \STATE \textbf{Return:} $(i^{*}, j^{*})$ and $I_{region}^{*}$.
    \end{algorithmic}
\end{algorithm}

Next, in the domain adaptation stage, the domain shift problem between two patches, which brings the different bright fields and device information, could make the noise for model training. Inspired by FDA\cite{yang2020fda}, the proposed framework conducts spectral transfer, mapping a benign sample to a malignant sample without changing semantic content. Given that ${F^A},\,{F^P}:{R^{H \times W \times 1}} \to {R^{H \times W \times 1}}$ are the amplitude and phase components of the Fourier transform $F$ of a mammogram patch, we have:

\begin{equation}
  F(x)(m,n) = \sum\limits_{h,w} {x(h,w){e^{ - k2\pi \left( {\frac{h}{H}m + \frac{w}{W}n} \right)}}},
\end{equation}
where $k^2 = -1$.

With mask $M_{\beta}$ contains zero value except for center region with $\beta \in (0,1)$ as follows:

\begin{equation}
  {M_\beta }(h,w) = {\Im _{(h,w) \in \left[ { - \beta H:\beta H, - \beta W:\beta W} \right]}},
\end{equation}
$\Im$ indicates an all-ones matrix. As shown in Fig.~\ref{fig: 2}, Benign patch and Malignant patch are $x^s \sim D^s$,$x^t \sim D^t$ respectively, FDA algorithm is shown as:

\begin{equation}
 \resizebox{.99\linewidth}{!}{$ {x^{s \to t}} = {F^{ - 1}}\left( {{M_\beta } \circ {F^A}({x^t}) + \left( {1 - {M_\beta }} \right) \circ {F^A}({x^s}),{F^P}({x^s})} \right) $},
\end{equation}
where $F^{-1}$ is the inverse Fourier transform mapping spectral information back to 2D-image space. The center (low frequency) part of the amplitude of the source image ${F^A}({x^s})$ will be transferred in the target style of ${x^t}$. This notation only modifies the amplitude component without altering the phase component ${F^P}$. Both components of the Fourier transform will be inversed back to a new image ${x^{s \to t}}$, whose remaining content of source image ${x^s}$ but the style of target image ${x^t}$.

Finally, the original malignant and domain-adapted benign are used for blending before pasting back to the benign sample. We proposed a novel soft mask and its inverse for mixing two patches. With any image having height $H$  and width $W$, a soft mask $S$ is defined as $S \in {\left[ {0,1} \right]^{H \times W}}$. Therefore, its inverse soft mask is $(1-S) \in {\left[ {0,1} \right]^{H \times W}}$. The output image mixing between two images $x^s, x^t$ is formularized as:

\begin{equation}
    \overline x  = \left( {S \otimes {x^t}} \right) \oplus \left( {\left( {1 - S} \right) \otimes {x^s}} \right), 
\end{equation}
% \[\overline x  = \left( {S \otimes {x^t}} \right) \oplus \left( {\left( {1 - S} \right) \otimes {x^s}} \right)\], 
whereas, $x^s, x^t$ are the source image (benign patch) and target image (malignant patch) respectively. The label of image $\overline x$ is the label of the target image.

The blending masks are generated following the Gaussian distribution. The gradient radial soft mask is the result of the outer product of two one-dimensional Gaussian distributions. It can be seen as:

\begin{equation}
    {S_W} = {e^{ - \frac{{{{\left( {x - {\mu _W}} \right)}^2}}}{{2{\sigma ^2}}}}},\,\,{S_H} = {e^{ - \frac{{{{\left( {x - {\mu _H}} \right)}^2}}}{{2{\sigma ^2}}}}},
\end{equation}
% \[{S_W} = {e^{ - \frac{{{{\left( {x - {\mu _W}} \right)}^2}}}{{2{\sigma ^2}}}}},\,\,{S_H} = {e^{ - \frac{{{{\left( {x - {\mu _H}} \right)}^2}}}{{2{\sigma ^2}}}}}\],
whereas $\mu _W$, $\mu _H$, and $\sigma$ are uniformly sampled from the input images' width W, height H ranges, and its spread in the image space, respectively. A sample of the mask can be seen in Fig.~\ref{fig: 2} and ~\ref{fig: 3}a.

\section{Experimental Settings}

\begin{table*}[t]
\caption{Quantitative results (\%) among our proposed DIVF frameworks, normal fusion frameworks, and no fusion approach}
\label{tab: divf}
\begin{center}
\begin{tabular}{l@{\hskip 0.1in}l@{\hskip 0.1in}c@{\hskip 0.1in}c@{\hskip 0.1in}c@{\hskip 0.1in}c}
    \toprule
    \textbf{Backbone}
     &
     & \multicolumn{2}{c} {\textbf{ResNet-18}}  
     & \multicolumn{2}{c} {\textbf{ResNet-34}}   \\
\toprule
Dataset & Method  &  F1-Score & AUC-ROC & F1-Score & AUC-ROC \\
\midrule
\multirow{5}{*}{VinDr-Mammo} & No Fusion & 70.12 & 68.79 & 71.48 & 70.22 \\
& Average Fusion & 72.54 & 74.20 & 73.25 & 72.88 \\
& Concatenate Fusion & 73.22 & 70.66 & 74.63 & 72.18 \\
& DIVF(Average) & 74.00 & 72.15 & 74.17 & 71.67 \\
& DIVF(Concatenate) & \textbf{75.34} & \textbf{74.24} & \textbf{75.98} & \textbf{74.86} \\
% VinDr-Mammo & No Fusion & 70.12 & 68.79 & 71.48 & 70.22 \\
% & Average Fusion & 72.54 & 74.20 & 73.25 & 72.88 \\
% & Concatenate Fusion & 73.22 & 70.66 & 74.63 & 72.18 \\
% & DIVF(Average) & 74.00 & 72.15 & 74.17 & 71.67 \\
% & DIVF(Concatenate) & \textbf{75.34} & \textbf{74.24} & \textbf{75.98} & \textbf{74.86} \\
\midrule
\multirow{5}{*}{CMMD} & No Fusion & 73.26 & 76.70 & 75.52 & 77.18 \\
& Average Fusion & 79.22 & 79.13 & 79.97 & 81.80 \\ 
& Concatenate Fusion & 75.86 & 77.10 & 78.12 & 77.67 \\
& DIVF(Average) & \textbf{81.45} & \textbf{84.14} & \textbf{82.44} & 80.92 \\
& DIVF(Concatenate) & 77.77 & 80.42 & 79.51 & \textbf{81.97} \\
\bottomrule
\end{tabular}
\end{center}
% \vskip -0.2in
\end{table*}

\subsection{Datasets}
\textbf{CMMD.} The Chinese Mammography Database (CMMD)\cite{cai2023online} includes 5.202 screening mammogram images conducted on 1.775 studies. We trained on 1.172 non-malignant mammograms and 2.728 malignant screening images with 85\%:15\% ratio splitting on the training set and test set. Furthermore, we employ stratified sampling, resulting in 498 benign and 1157 malignancy ipsilateral view samples on the training set and 88 benign and 205 malignancy ipsilateral view samples on the testing set.

\vspace{3mm}\hspace{-5mm}\textbf{VinDr-Mammo.} A large-scale full-field digital mammography dataset\cite{nguyen2023vindr}, which contains 20.000 scans from 5.000 studies of  Vietnamese patients. Because of the untrustworthy of BI-RADS 3, the inconsistency between BI-RADS 4 and 5, and the heavy imbalance of BI-RADS 1, we arrange the image-level labels into two classes: Suspicious Benign (BI-RADS 2) and Suspicious Malignancy (BI-RADS 4 and 5). So as the preprocessing on CMMD, there are 2.831 ipsilateral view samples (CC and MLO views on the same breast) that were split into training set (1870 benign and 395 malignancy cases) and testing set (467 benign and 99 malignancy cases). Besides, for the malignant lesions synthesis framework, we use all region-level annotations for making new malignant samples.

\subsection{Implementation Details}
ResNet family architectures are used for the Feature Extractor part of the framework, including ResNet-18 and ResNet-34. In the data loading part, the images are loaded with a batch size of 32 (two ipsilateral views for each breast with a total of 16 breasts). The model was trained for 200 epochs using SGD optimizer\cite{ketkar2017stochastic} with an initial learning rate $1 \times 10^{−3}$ and decays by $0.1$ after $20, 40, 60$, and $80$ epochs. We resized images to the same 800 x 800 for both the training and testing phases. Our work was built on Pytorch version 1.9.1 and trained by using NVIDIA RTX 3090Ti GPU (24GB). We used the Macro F1-Score to evaluate and reduce the imbalance effectiveness in the dataset, which is computed using the arithmetic (unweighted) mean of all the per-class F1 scores. Besides, the Area under the ROC Curve (ROC AUC) is used for measuring the model performance under slightly imbalanced dataset training. 

\section{Results and Ablation Studies}
\subsection{Dual Ipsilateral Views Fusion Network}
In this section, there are three main approaches we want to test. 1) No Fusion, a single view is fed into the backbone, no combining of two views CC and MLO in this case. 2) Average Fusion and Concatenate Fusion, there is no skip connection with two examined features and fused features in the Features Fusion Blocks phase. 3) DIVF, contains all components described in Section \ref{divf}. Table~\ref{tab: divf} shows the testing results of our proposed methods on VinDr-Mammo and CMMD datasets. Our DIVF framework shows a significant improvement compared to the conventional techniques, with a mean of around 5\% on VinDr-Mammo and 7\% on CMMD. For each method of combining features, the DIVF shows the apparent effectiveness of the feature fusion mechanism for classifying the benign and malignant. Testing on VinDr-Mammo, DIVF Framework with concatenate method achieves the highest Macro F1-score and AUC-ROC on both backbones, 75.98\% and 74.86\% respectively. Different from VinDr-Mammo, we use average aggregation with the DIVF method seems to be more robust on CMMD. This strategy outperforms the normal fusion or no fusion approaches, which achieved 81.45\% on ResNet-18 and 82.44\% on ResNet-34 in Macro F1-score evaluation metrics.

\begin{figure}[ht]
	\centering
	\includegraphics[width=0.85\linewidth]{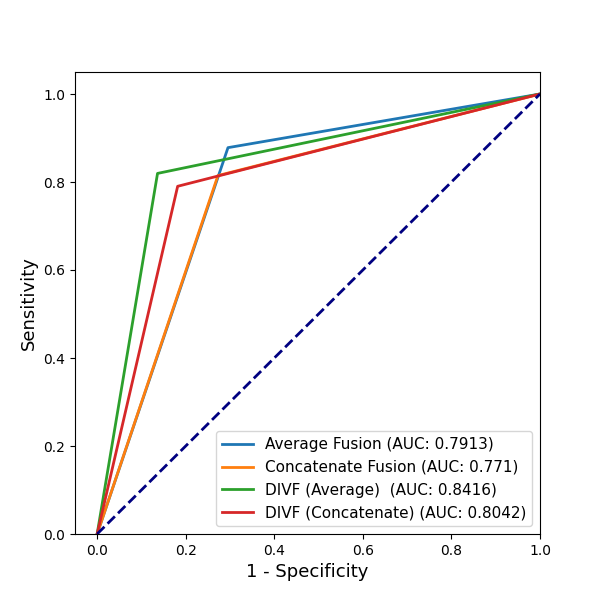}
	\caption{AUC-ROC for benign/malignant classification on CMMD dataset. Testing performance of the average fusion, concatenate fusion, EA average fusion, and EA concatenate fusion}
	\label{fig: aucroc}
\end{figure}

Fig.~\ref{fig: aucroc} highlights the trade-off between the true positive rate (sensitivity) and the false positive rate (1-specificity) by plotting the ROC curve for malignant and benign categorization. The DIVF (Average) achieved the best performance with a sensitivity of 87.8\% and a specificity of 70.45\%, which resulted in 0.8416 of AUC. Continuously, the second high performance is the DIVF (Concatenate), which obtained 80.42\%, lower than 3.74\% compared with the best one. In contrast, Average Fusion and Concatenate Fusion do not outcome the DIVF version which achieved 79.13\% and 77.1\%, respectively. This overcome can be explained in the way we support the model by adding the examined features in the features fusion blocks phase of the framework. After going through the IVF block, fused features might lose detailed information because fusion operation tends to generalize the feature in both views. Thus, this alleviates the examined features. Therefore, adding the examined features prevents two problems: solving the vanishing problem in the IVF block and diversifying information.

\subsection{Soft-Adapted Malignancy Synthesis Framework}

\begin{table*}[t]
\caption{Ablation studies of our proposed Soft-Adapted Malignancy Synthesis Framework on DIVF Concatenate with ResNet-34 on VinDr-Mammo Dataset}
\label{tab: ablation studies}
\begin{center}
\begin{tabular}{l@{\hskip 0.2in}|c@{\hskip 0.2in}c@{\hskip 0.1in}c@{\hskip 0.1in}c@{\hskip 0.1in}|c@{\hskip 0.1in}c}

\toprule
 & DIVF & \makecell{Region Selection \\ \& CutMix}  & Fourier Adaptation & Soft Mask & Macro F1-Score \\
\midrule
\midrule
Baseline & \checkmark&  &  &  &  75.98\\
\midrule
\multirow{5}{*}{Proposed Methods} & \checkmark& \checkmark  &  &  &  76.54 \\ \cmidrule(){2-7}
     & \checkmark& \checkmark  & \checkmark &  &  76.96\\ \cmidrule(){2-7} 
     & \checkmark& \checkmark  &  & \checkmark &  76.78 \\ \cmidrule(){2-7}
     & \checkmark& \checkmark  & \checkmark & \checkmark &  \textbf{77.32}& \\
\bottomrule
\end{tabular}
\end{center}
% \vskip -0.2in
\end{table*}

Table~\ref{tab: ablation studies} shows the ablation studies of our proposed synthesis framework on VinDr-Mammo malignant sample synthesis. As shown in the table, we can see the effect of each element contributing to the final F1 score of our method. The whole framework combined three mechanisms for creating new samples achieves 77.02\% on the F1-Score metric. The limitation of the original CutMix seems to be eliminated with Fourier Adaptation and Soft Mask. The new samples are no longer containing bad-looking malignant tumors with different color-style and hard contours when conducting copy-and-paste patches. The detailed outputs of each part in our framework are visualized in Fig.~\ref{fig: 3}b-f.

\begin{figure}[ht]
	\centering
	\includegraphics[width=0.8\linewidth]{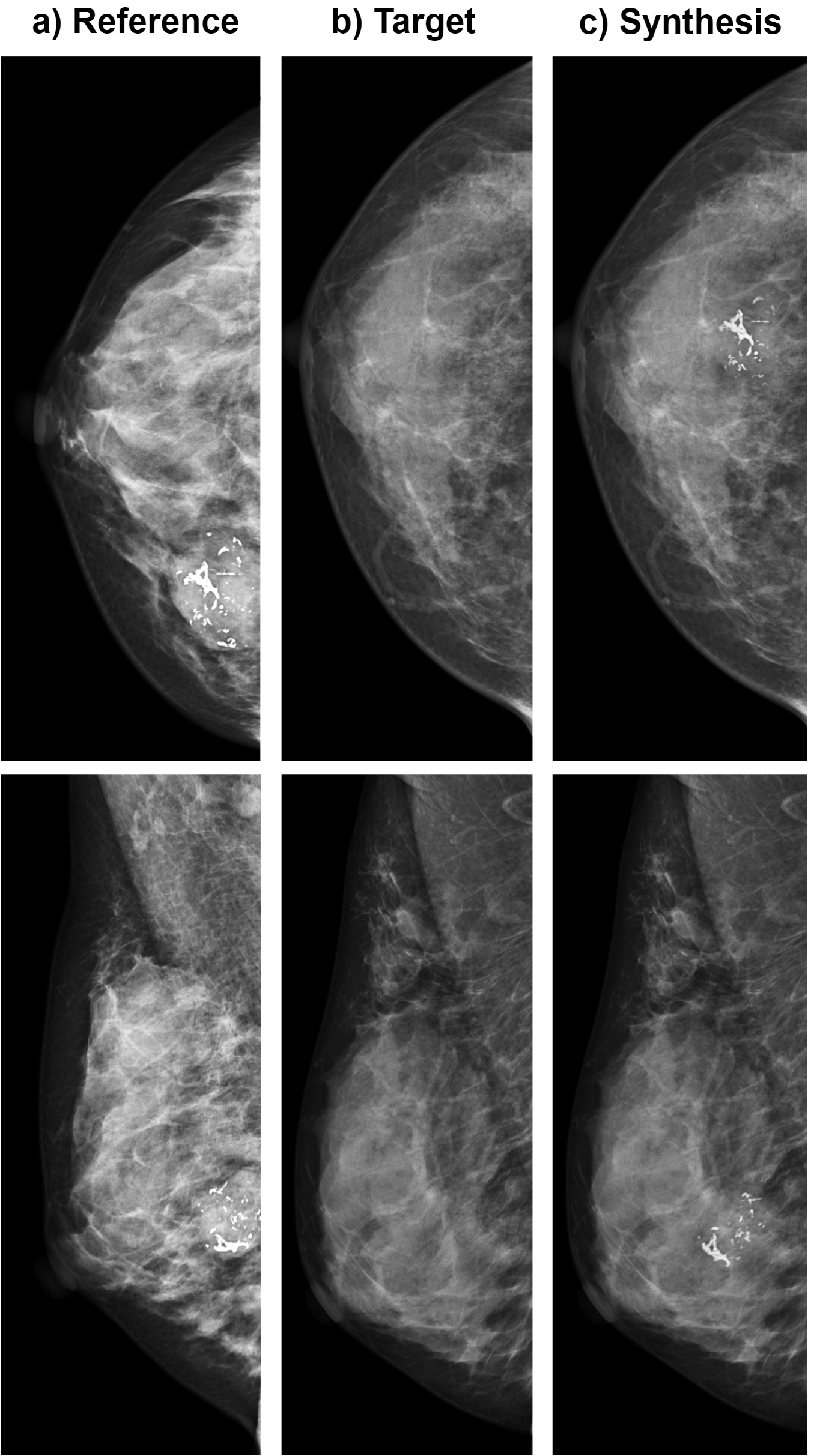}
	\caption{Our framework results on CC (top) and MLO (bottom) views. a) Reference image b) Target image c) Synthesis image.}
	\label{fig: calc}
\end{figure}

Fig.~\ref{fig: 3}d shows the synthesis image using the CutMix algorithm. In the replaced region, the cancer mass seems incompatible with the source image in the style. This can cause the unwanted detect edge in CNN sliding filters, thus leading to outlier features and poor representation. This strategy achieved a slight improvement in performance with (+0.56\%) compared with the baseline model, ResNet-34 DIVF Concatenate in Table~\ref{tab: divf}. Furthermore, the results also increase a bit (+0.42\%) when the Fourier Domain Adaptation method is applied. Fig.~\ref{fig: 3}e proves the improvement with smooth style in the replaced region. However, the suddenly changing pixel value that occurs on the top-left corner of the transformed region does not perfectly make the synthesis image look natural. Afterward, our proposed Soft-Adapted Malignancy Synthesis Framework can alleviate those problems which perfectly adapting the target style to the source image. Fig.~\ref{fig: 3}f and Fig.~\ref{fig: calc}c show natural-looking, yet trustworthy, mammography screening that achieved 77.32\% on Macro F1-Score. Those upsampling data shown in Fig.~\ref{fig: calc}c, created by Fig.~\ref{fig: calc}a,b, are reliable for the training stage to handle most of the imbalance mammogram dataset. This framework has shown its robustness on many different types of lesions including Mass, Calcification, Asymmetry, etc.

%Fig.~\ref{fig: 3}f looks like a real mammography screening that outperformed previous works that achieved 77.32\% on Macro F1-Score.

%We visualize more examples as shown in Fig.~\ref{fig: calc} on CC and MLO views. Those samples also look real that can become training data to reduce the imbalance effect in dataset. 

\section{Conclusion}

%In this work, we proposed a DIVF framework to leverage the ipsilateral multi-view information for classifying cancerous mammograms. The model learns the low-level features separately from two ipsilateral views and conducts feature aggregation for fusion learning on the high-level features. The proposed IVF block takes both examined view and the auxiliary view for features fusion to enhance the examined view. Besides, we also proposed a new natural-looking malignant lesions synthesis framework for creating new reliable samples to improve the classification results. The simple framework, yet novelty, helps to eliminate the biases when conducting CutMix with style adaptation and smooth gradient contour of the lesion area. From our experimental setting, our proposed methods achieve significant improvement compared to other conventional methods. We achieved the SOTA breast cancer classification performance and generalizability across two datasets. In the future, we will expand the task into lesion detection or density classification. We will also run more statistical analyses to reveal the meaning behind both contributions' concepts.

In this work, we proposed a DIVF framework to leverage the ipsilateral multi-view information for classifying cancerous mammograms. The model learns the low-level features separately from two ipsilateral views and conducts feature aggregation for fusion learning on the high-level features. Our model learned low-level features from two ipsilateral views and effectively fused high-level features. The IVF block enhanced the examined view, resulting in improved classification. Additionally, our natural-looking malignant lesions synthesis framework generated reliable samples, leading to state-of-the-art performance and generalizability across two datasets. Our research shows promise for enhancing breast cancer diagnosis and treatment. Future work aims to extend our research to lesion detection and density classification tasks and conduct further statistical analyses to gain deeper insights.

\section{Acknowledgement}
This paper is partially supported by AI VIETNAM. We thank Integrated MechanoBioSystems Lab (IMBSL) from the Biomedical Engineering Department of National Cheng Kung University for providing the GPU to support the numerical calculations in this paper.

{\small
\bibliographystyle{ieee_fullname}
\bibliography{egbib}
}

\end{document}